\documentclass{Interspeech}

\usepackage{amsmath,amssymb}
\usepackage[table]{xcolor}
\usepackage{multirow}
\RequirePackage{booktabs}
\usepackage{bbm}
\usepackage{cite}
\usepackage{color} 
\usepackage{soul} 

\interspeechcameraready

\title{Improving Synthetic Data Training for Contextual Biasing Models with a Keyword-Aware Cost Function}

\author[affiliation={1}]{Chin Yuen}{Kwok}
\author[affiliation={1}]{Jia Qi}{Yip}
\author[affiliation={1}]{Eng Siong}{Chng}

\affiliation{College of Computing and Data Science}{Nanyang Technological University}{Singapore}
\email{kwok0062@e.ntu.edu.sg}

\usepackage{comment}

\begin{document}

\maketitle

\begin{abstract}

     Rare word recognition can be improved by adapting ASR models to synthetic data that includes these words. Further improvements can be achieved through contextual biasing, which trains and adds a biasing module into the model architecture to prioritize rare words. While training the module on synthetic rare word data is more effective than using non-rare-word data, it can lead to overfitting due to artifacts in the synthetic audio. To address this, we enhance the TCPGen-based contextual biasing approach and propose a keyword-aware loss function that additionally focuses on biased words when training biasing modules. This loss includes a masked cross-entropy term for biased word prediction and a binary classification term for detecting biased word positions. These two terms complementarily support the decoding of biased words during inference. By adapting Whisper to 10 hours of synthetic data, our method reduced the word error rate on the NSC Part 2 test set from 29.71\% to 11.81\%.
\end{abstract}

\section{Introduction}

Accurate recognition of rare words in automatic speech recognition (ASR) is crucial, as these words often carry key semantic information. However, ASR models frequently misidentify them due to their limited presence in training data. Contextual biasing addresses this challenge by incorporating additional contextual information, such as contact lists, domain-specific terms, or user-defined vocabularies, to improve recognition. This technique is typically implemented through specialized modules \cite{xu2023adaptive,harding23_interspeech,sathyendra2022contextual,naowarat23_interspeech,shamsian2024keyword,sun23e_interspeech,futami2024phoneme} integrated into pre-trained ASR models, which adjust model activations based on a predefined list of biasing words, thereby enhancing recognition accuracy.

\begin{figure}[t]
  \centering
  \includegraphics[width=\linewidth]{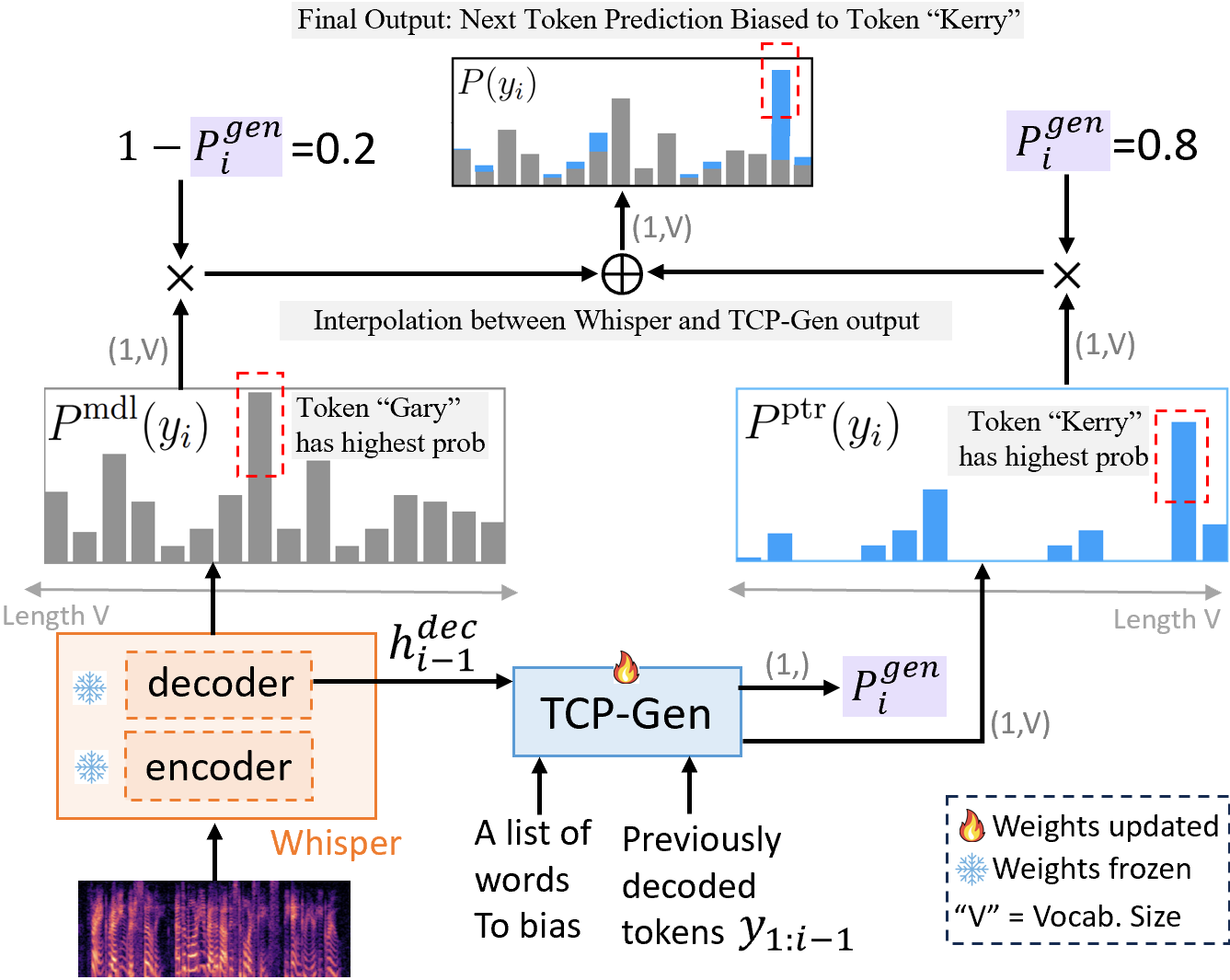}
  \caption{Example of applying TCPGen to Whisper for contextual biasing to predict the $i$-th token. Tensor shapes are shown in parenthesis. Bottom-left: Whisper takes in an audio to output next token probabilities $P^{mdl}(y_{i})$. Bottom-center: TCPGen takes in Whisper decoder's last hidden state $h^{dec}_{i-1}$, a list of words to bias and the previously decoded tokens $y_{1:i-1}$ to output biased probabilities $P^{ptr}(y_{i})$ and an interpolation weight $P^{gen}_i$. Top-center: Interpolation between $P^{mdl}(y_{i})$ and $P^{ptr}(y_{i})$ is performed using the weight $P^{gen}_i$ to output the final probabilities $P(y_i)$.
}
  \label{fig:whisper_tcpgen_architecture}
\end{figure}

While training these modules on non-rare-word data improve rare word recognition, they remain less effective than direct training on data containing these rare words, which is often unavailable. This data limitation creates a bottleneck for effective contextual biasing. To address this, synthetic training data generated via text-to-speech (TTS) systems can be used as a substitute \cite{Zheng2020UsingSA}. By generating audio samples from sentences containing the target rare words, TTS enables the creation of training datasets tailored to specific contexts. This approach alleviates the need for extensive real-world recordings and allows the ASR model to automatically learn rare words.

However, synthetic audio data introduce artifacts that differ from natural speech, such as unnatural prosody or acoustic distortions, which can lead to overfitting when training the biasing module. Modules trained on such data may perform poorly when exposed to natural audio, limiting the effectiveness of contextual biasing in real-world scenarios. To mitigate overfitting, prior work has introduced training constraints to the standard ASR loss used for training the biasing modules, addressing two key challenges in contextual biasing:
\begin{enumerate}
\item \textbf{When to perform biasing}: Xu et al. \cite{xu2023adaptive} introduced an Entity Detector to identify the presence of bias words, enabling selective application of biasing. Similarly, Huang et al. \cite{huang2023spike} refined biasing by modifying layer outputs only at time frames corresponding to bias phrases.
\item \textbf{What to bias}: Kulshreshtha et al. \cite{kulshreshtha2023multilingual} restricted the CTC encoder to cross-attend only to the correct bias phrase from a given list. Tang et al. \cite{tang2024improving} extended this by constraining the prediction network.
\end{enumerate}

Although these approaches are effective, they rely on assumptions that may not hold for speech foundation models like Whisper \cite{Radford2022RobustSR}. Specifically, they assume that encoders can produce frame-level predictions \cite{kwok2024improved} or that audio and text modalities are modeled separately through distinct encoder and predictor networks. However, Whisper does not generate frame-level predictions, and its decoder processes both encoded acoustic features and text-based inputs via cross- and self-attention, challenging the applicability of these methods to such architectures.

\begin{figure}[t]
  \centering
  \includegraphics[width=0.95\linewidth]{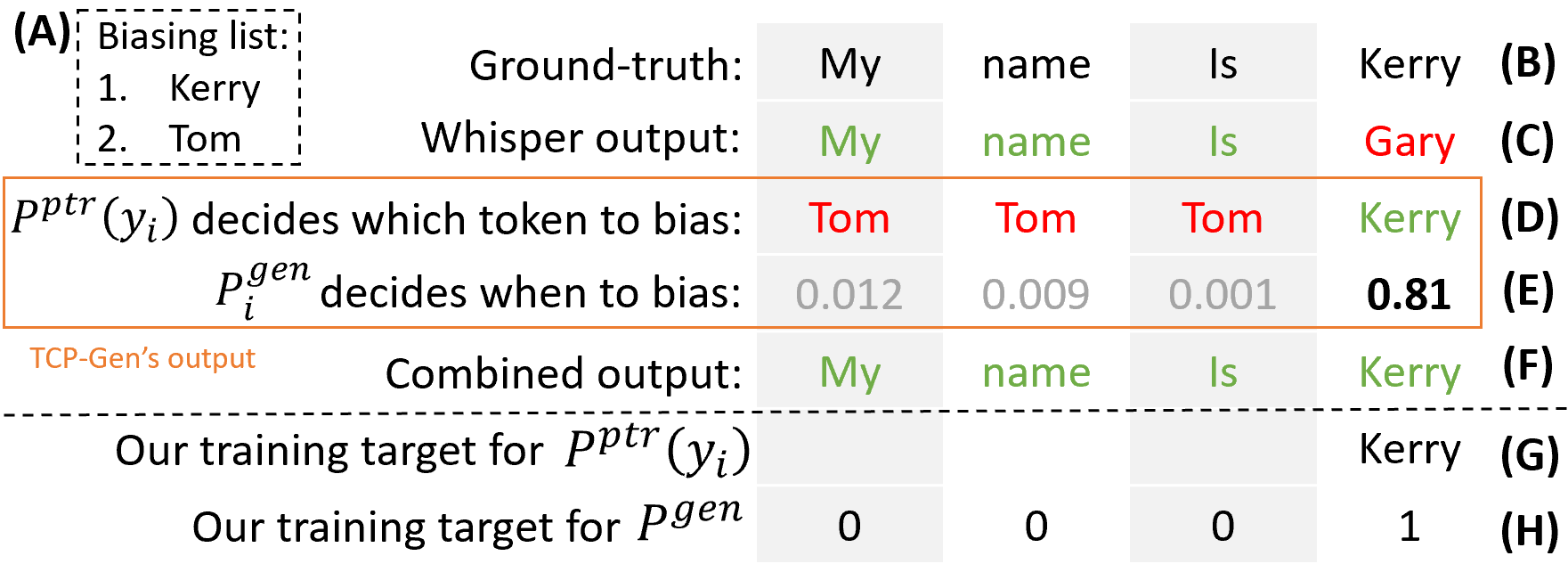}
  \caption{Overview of our modified TCPGen training objective. As an example, (B) given an audio with ground truth transcript ``My name is Kerry", (C) Whisper wrongly outputs ``Gary" instead. To perform contextual biasing, (A) a biasing list is provided to TCPGen. (D) Then it predicts the token to bias at each transcript position and (E) if biasing should be performed. Originally, (F) the combined output of Qwen Audio and TCPGen is trained on the vanilla ASR cross-entropy loss using the ground-truth in (B) as training targets. To better guide the TCPGen to model the pronunciations of rare words, we propose to replace the ASR loss with two loss that explicitly trains TCPGen to output the desirable (G) $P^{ptr}(y_i)$ and (H) $P^{gen}_i$ targets.
}
  \label{fig:bias}
  \vspace{-5 pt}
\end{figure}

To address this, we integrate the pointer-network-based biasing module TCPGen \cite{sun23e_interspeech} into Whisper for decoder-side adaptation \cite{kwok2024continual,kwok2024continual_2}. We propose to modify the ASR objective function, which is often used to train the biasing modules, to additionally focus on the biasing words to reduce overfitting. Specifically, given that TCPGen is trained to output a bias probability $P^{ptr}(y_i)$ and an interpolation weight $P^{gen}_i$ as shown in Fig. \ref{fig:whisper_tcpgen_architecture}, we introduce two novel loss terms: 1) to determine what to bias, we introduce a masked cross-entropy loss for $P^{ptr}(y_i)$ to predict the posterior probabilities of the rare words as shown in Fig. \ref{fig:bias}G. 2) to determine when to perform biasing, we introduce a binary classification for $P^{gen}_i$ to indicate the presence of biasing words at each token position, as shown in Fig. \ref{fig:bias}H. These two objectives complementarily support the decoding of rare words during inference.

Our contributions are fourfold. We show that: 1) applying the vanilla TCPGen module to Whisper can cause overfitting, 2) by changing the objective function of TCPGen to our two novel loss terms, a significant improvement of up to 16.6\% relative WER reduction is observed. 3) We show that our two novel loss terms, which are specifically designed for contextual biasing, can be used without the standard ASR loss to effectively train the biasing modules, unlike previous work that still includes the ASR loss when training the modules. 4) Our objective function improves the interpretability of the TCPGen output $P^{gen}_i$ and allows explicit analysis of the false acceptance rate for contextual biasing.

\section{Method}
\label{sec:method}

\subsection{Tree-constrained pointer generator (TCPGen)}

TCPGen is a neural network component that is added to ASR models for contextual biasing, where TCPGen is trained end-to-end and the weights of the pre-trained ASR model is frozen. As shown in Fig. \ref{fig:whisper_tcpgen_architecture}, at output step $i$, TCPGen takes the last hidden state $h^{dec}_{i-1}$ of Whisper's decoder, a list of words to bias and the previously decoded tokens $y_{1:i-1}$ to output a distribution $P^{ptr}(y_i)$ over all tokens to determine which token to bias. Furthermore, TCPGen predicts a generation probability $P^{gen}_i$ that determines when to perform biasing. The final output $P(y_i)$ is an interpolation between $P^{ptr}(y_i)$ and the original ASR model distribution $P^{mdl}(y_i)$ weighted by $P^{gen}_i$:

\begin{equation}
P(y_i) = P^{mdl}(y_i)(1 - P^{gen}_i) + P^{ptr}(y_i) P^{gen}_i
\label{eq:final_output}
\end{equation}

An example of contextual biasing is shown in Fig. \ref{fig:bias}, where given an audio with ground truth transcript ``My name is Kerry", Whisper incorrectly outputs ``Gary" instead of ``Kerry". To perform contextual biasing, a biasing list consisting of ``Kerry" and ``Tom" is provided to TCPGen. Then TCPGen outputs $P^{ptr}(y_i)$ at each decoding step $i$ to find the most probable word to bias as shown in Fig. \ref{fig:bias}D, and uses $P^{gen}_i$ to decide when to perform biasing as shown in Fig. \ref{fig:bias}E. Finally, the combined output of whisper and TCP-Gen has the output ``Gary" biased to ``Kerry" as shown in Fig. \ref{fig:bias}F.

\subsection{Training Objectives}
\label{sec:loss}

Originally, TCPGen is trained to optimize the vanilla ASR cross entropy loss $\ell_{asr}$:

\begin{equation}
\ell_{asr} = -\sum_{i=1}^U \sum_{c=1}^V \mathbbm{1}(y_i=c)\log(P(y_i=c))
\label{eq:asr}
\end{equation}

where $U$ and $V$ are the target sequence length and vocabulary size and $\mathbbm{1}(y_i=c)$ is an indicator function that outputs $1$ if the $i$-th token in $y$ is token $c$ or outputs $0$ otherwise.

\subsubsection{Binary Classification Loss}

To more effectively train TCPGen and to reduce overfitting, we propose replacing the ASR loss of TCPGen with two losses $\ell_{ptr}$ and $\ell_{gen}$ to explicitly guide TCPGen to learn the desired $P^{ptr}(y_i)$ and $P^{gen}_i$ outputs, respectively. Specifically, as shown in Fig. \ref{fig:bias}H, $P^{gen}_i$ should be explicitly optimized to classify if the $i$-th decoded token needs biasing:

\begin{multline}
\ell_{gen} = -\sum_{i=1}^U \alpha \cdot \mathbbm{1}(i\in K) \log(P^{gen}_i)\\ + (1-\alpha) (1-\mathbbm{1}(i\in K)) \log(1-P^{gen}_i)
\label{eq:loss_gen}
\end{multline}

where $K$ is the set of the position indexes of the biased tokens, e.g. the index of ``Kerry" is $4$ in Fig. \ref{fig:bias}B, and $P^{gen}_i$ is trained on the negative log-likelihood loss $\ell_{gen}$ to perform binary classification. 

As rare and non-rare word distributions are imbalanced, we include a weight $\alpha>0.5$ to further encourage the model to classify the minority class.

\subsubsection{Masked Cross Entropy Loss}

Next, as shown in Fig. \ref{fig:bias}G, $P^{ptr}(y_i)$ should be optimized to predict the rare word probabilities only for tokens that need biasing:

\begin{equation}
\ell_{ptr} = -\sum_{i=1}^U \sum_{c=1}^V \mathbbm{1}(i\in K) \mathbbm{1}(y_i=c)\log(P^{ptr}(y_i=c))
\end{equation}

This is different from optimizing $\ell_{asr}$ as $P^{ptr}(y_i)$ will redundantly learn to predict the rare word probabilities at the positions where there are no rare words under the formulation of $\ell_{asr}$ in Eq. \ref{eq:asr}, which leads to overfitting. Given the two losses, TCPGen is trained to optimize $\ell_{gen} + \ell_{ptr}$.

\subsection{False Acceptance Rate (FAR)}
\label{sec:far}

To analyze the effectiveness of our proposed objective functions, the false acceptance rate (FAR) and the true acceptance rate (TAR) are calculated for $P^{gen}_i$. False acceptance occurs when the $i$-th decoded token needs biasing, but $P^{gen}_i$ predicts otherwise. True acceptance occurs when the $i$-th decoded token needs biasing and $P^{gen}_i$ predicts correctly.

We emphasize that this analysis is only possible with our $\ell_{gen}$ formulation. This is because in the original $\ell_{asr}$ formulation, $P^{gen}_i$ is optimized as an interpolation weight in the range of $(0,1)$ and a higher value means that biasing is needed for the $i$-th decoded token, but the decision boundary is not obvious. By explicitly optimizing $P^{gen}_i$ to perform binary classification using our $\ell_{gen}$, the FAR and TAR can be calculated by setting the decision boundary to $0.5$.

\subsection{Masked Probabilities}
\label{sec:mask_prob}

Given that $M_i$ is the set of tokens to bias at the $i$-th decoding step. TCPGen originally assigns zero probabilities to $P^{ptr}(y_i=c):c \not\in M_i$ such that according to Eq. \ref{eq:final_output}:

\begin{equation}
 \forall c \not\in M_i:P(y_i=c) = P^{mdl}(y_i=c) (1-P^{gen}_i)
\label{eq:mask_p}
\end{equation}

where $P(y_i=c):c \not\in M_i$ is just a scaled version of $P^{mdl}(y_i=c)$. However, we empirically find that the scaling effect may be undesirable when we optimize $P^{gen}_i$ to have values close to $1$ using $\ell_{gen}$ as it completely suppresses $P(y_i=c):c \not\in M_i$ and increases the chance of overbiasing. Therefore, we remove the $(1-P^{gen}_i)$ term in Eq. \ref{eq:mask_p} and set $\forall c \not\in M_i:P(y_i=c) = P^{mdl}(y_i=c)$.

\section{Experiment Setup}
\label{sec:exp}

\subsection{Dataset and metrics}
\label{ssec:dataset}

Experiments were conducted on the National Speech Corpus Part 2 (NSC-Part-2) dataset \cite{koh2019building}, a subset of a large-scale Singapore English corpus that includes road names and addresses. They are recordings of people asking for directions and consist of 13K unique utterances. This dataset is chosen because synthetic audio has been publicly released for the dataset \cite{yuen2023asr}, and about $15\%$ of the words in the synthetic audio are testset rarewords, which is significantly more than Librispeech\footnote{Although synthetic audio has also been released for Librispeech \cite{zen2019libritts}, only about $2\%$ of the words are testset rarewords. As a result, synthetic data training for Librispeech is less effective.}.

We use the 20 hours of real audio train set, 10 hours for synthetic audio train set (VITS-SPKSET1 \cite{yuen2023asr}) and 2 hours of real audio test set for our experiments. Both the synthetic and real audio train sets used contain the same number of sentences. Synthetic audios are generated from the same text as the real train set, using the VITS \cite{kim2021conditional} text-to-speech model from coqui-ai\footnote{\url{https://github.com/coqui-ai/TTS}}, which is trained on the VCTK corpus \cite{Yamagishi2019CSTRVC}. We find that although the synthetic audio has a different English accent compared to the NSC-Part-2 dataset, it is sufficient to adapt Whisper to the NSC-Part-2 text domain using the audio. 

Similar to Sun et al. \cite{sun2023contextualbiasingremaineffective}, we treat words that do not exist in the 10K common word list \footnote{https://github.com/first20hours/google-10000-english/blob/master/google-10000-english.txt} as rare words. In total, there are 3856 and 1106 rare words in the train and test sets respectively. In this setup all the testset rarewords are considered OOV words as we never train TCPGen with any real audio data containing the words. During inference, a list of biasing words for each utterance was obtained by collecting rarewords that appear in the utterance and adding $N$ distractors, which are randomly sampled from the set of all possible rarewords.

In addition, we also experiment on DSTC2 \cite{henderson-etal-2014-second}, a human-machine task-oriented dialogue dataset where each of 10 hours of user-side speech input was used for training and evaluation. The biasing list for inference was organised by the error-based biasing list obtained from the training set following \cite{sun2023contextualbiasingremaineffective}. The full biasing list contains 274 words.

To evaluate our method, two additional metrics were used to evaluate system performance besides word error rate (WER): the biased word error rate (B-WER) and the unbiased word error rate (U-WER) \cite{le2021contextualized}. B-WER is calculated as the number of error words from the biasing list divided by the total number of words from the biasing list in the test set, with insertion errors counted if the inserted word is from the bias list. U-WER is similar to B-WER but focuses solely on words not belonging to the biasing list.

\subsection{Implementation details}
\label{ssec:implementation}

We implement our methods based on the popular SpeechBrain \cite{speechbrain} toolkit. We adapt whisper-small and Qwen Audio with vanilla fine-tuning (FT) or AGEM \cite{lopez2017gradient} for 2 epochs and set the train batch size to 6. AGEM is a regularization method that constrains the gradients, and we find it to be effective in reducing overfitting on synthetic training data. We simply follow the setup from Kwok et al. \cite{kwok2024continual} to apply AGEM. We set the learning rate to $0.005$ and $0.0001$ for Whisper and Qwen Audio respectively. We use the AdamW optimizer with a variant\footnote{\url{https://speechbrain.readthedocs.io/en/latest/\_modules/speechbrain/nnet/schedulers.html\#NewBobScheduler}} of the ReduceLROnPlateau learning rate (LR) scheduler. The encoder weights are frozen to prevent overfitting \cite{kwok2024low,kwok2025extending}. We sweep through the hyper-parameters to optimize WER.

\section{Results and Discussions}
\label{sec:results}

\begin{table*}[t]
  \centering
  \caption{\small OOV contextual biasing for whisper-small. Either the real of synthetic (Syn) version of the NSC-Part-2 train/development set is used for training/hyper-parameter tuning. $N$ is the number of distractors added to the biasing words list. The first block of results have repeated results as $N$ is irrelevant. TCPGen-2L means TCPGen trained with our two novel loss terms.}
  \label{tab:exp-librispeech}
  \makebox[\textwidth][c]{
  \begin{tabular}{lcccccccccccc} 
    \toprule
    \multirow{2}{3.5cm}{Method} & \multicolumn{2}{c}{NSC-Part-2} & \multicolumn{3}{c}{N=10} & \multicolumn{3}{c}{N=50} & \multicolumn{3}{c}{N=100} \\ 
    \cmidrule(lr){2-3} \cmidrule(lr){4-6} \cmidrule(lr){7-9} \cmidrule(lr){10-12} 
    & Real & Syn & WER & BWER & UWER & WER & BWER & UWER & WER & BWER & UWER \\
    \midrule
    unadapted & & & 29.71 & 54.03 & 17.21 & 29.71 & 54.03 & 17.21 & 29.71 & 54.03 & 17.21 \\
    Kwok et al. \cite{yuen2023asr} & & \checkmark & 16.5 & - & - & 16.5 & - & - & 16.5 & - & - \\
    FT & \checkmark & & 9.40 & 16.95 & 5.52 & 9.40 & 16.95 & 5.52 & 9.40 & 16.95 & 5.52 \\ 
    FT & & \checkmark & 15.00 & 28.02 & 8.31 & 15.00 & 28.02 & 8.31 & 15.00 & 28.02 & 8.31 \\
    AGEM \cite{lopez2017gradient} & & \checkmark & 14.16 & 28.80 & 6.64 & 14.16 & 28.80 & 6.64 & 14.16 & 28.80 & 6.64 \\
    \cmidrule{1-12}
    \multicolumn{1}{l}{\textit{reproduced baselines}}\\
    AGEM \cite{lopez2017gradient} & & \checkmark &\\
    \multicolumn{2}{l}{\quad + Contextual Adapter \cite{gong2024contextual}} & \checkmark & 15.90 & 31.11 & 8.08 & 15.72 & 30.71 & 8.01 & 15.76 & 30.68 & 8.09 \\
    \quad + TCPGen \cite{sun2023contextualbiasingremaineffective} & & \checkmark & 14.16 & 28.80 & \underline{6.64} & 14.16 & 28.80 & \underline{6.64} & 14.16 & 28.80 & \underline{6.64} \\
    \quad + Alter. pron. \cite{le2020g2g} &  & \checkmark & 14.20 & 28.29 & 6.96 & 14.54 & 28.29 & 7.47 & 15.01 & 28.45 & 8.11\\
    \quad + Alter. pron. \cite{le2021deep} & \checkmark &  & 13.08 & 25.39 & 6.76 & 13.75 & 26.28 & 7.32 & 14.56 & 27.32 & 8.00\\
    \cmidrule{1-12}
    \multicolumn{1}{l}{\textit{our method}}\\
    AGEM & & \checkmark & \\
    \quad + TCPGen-2L & & \checkmark & \textbf{11.81} & \textbf{20.74} & 7.22 & \textbf{12.81} & \textbf{22.65} & 7.75 & \textbf{12.86} & \textbf{22.41} & 7.95 \\
    \bottomrule
  \end{tabular}
  }
\end{table*}

\begin{table}[t!]
    \caption{Contextual biasing results ($N=100$) for Qwen-Audio on the DSTC2 dataset. $\Delta$ BWER means relative \% improvement in BWER. TCPGen-2L means TCPGen trained with our two novel loss terms.} 
    \centering
    \label{tab:main2}
    \begin{tabular}{lcccc}
    \toprule
    Method & WER & \scalebox{.9}[1.0]{BWER} & \scalebox{.9}[1.0]{UWER} & \scalebox{.9}[1.0]{$\Delta$BWER} \\
    \hline
    unadapted & 26.5 & 107 & 25.2  & - \\
    Sun et al. \cite{sun2023contextualbiasingremaineffective} & 13.9 & & & \\
    FT & 7.39 & 87.9 &6.14 &  0.0 \\
    \quad + TCPGen & 7.39 & 87.9 &6.14 &  0.0 \\
    \quad + TCPGen-2L & \textbf{7.01} & \textbf{72.9} &\textbf{5.98} &  \textbf{17.1} \\
    \bottomrule
\end{tabular}
\vspace{-0.2cm}
\end{table}

The contextual biasing results of whisper-small are shown in Table \ref{tab:exp-librispeech}. The first block of the table shows that whisper-small originally performs worse on NSC-Part-2, and the performance improves if the model is vanilla fine-tuned (FT) on the synthetic data. AGEM further improves the performance, as it regularizes the training on synthetic data.

Based on the AGEM model, we further apply contextual biasing baseline methods as shown in the second block of the table. First, the results of contextual adapters and TCPGen show no improvements, and we observe that TCPGen's $P^{ptr}(y_i)$ is always close to $0$. We hypothesize that this occurs because the AGEM model is already overfitted to the NSC-Part-2 synthetic train set and has been fully optimized with respect to the ASR loss. Consequently, further training of the contextual biasing modules, which are essentially error correction modules, can barely optimize the ASR loss as the AGEM model has almost no errors on the train set already. Instead, further training only overfits the modules to the artificial patterns of the synthetic audio.

Next, when we apply our two novel losses onto TCPGen (TCPGen-2L), significant improvement is observed over the vanilla TCPGen as we hypothesize that our method can effectively optimize the contextual-biasing-specific objectives without relying on the ASR loss and this reduces overfitting.

\subsection{More overfitting scenarios}

To validate the effectiveness of our method, we further perform experiments using Qwen Audio on the DSTC2 dataset. Specifically, we first fine-tune (FT) Qwen Audio on DSTC2 real train set, and further train TCPGen on top of the fine-tuned model using the same train set. 

The results in Table \ref{tab:main2} show that originally the unadapted Qwen-Audio has a large BWER as the biasing list is error-based -- the words are only included in the list if they are originally wrongly recognized by Qwen-Audio. Then, vanilla TCPGen show no improvements, and we observe that TCPGen's $P^{ptr}(y_i)$ is always close to $0$. This is expected, as we hypothesize that the FT model is already fine-tuned to the DSTC2 real train set and the ASR loss is already optimized. As a result, further training the contextual biasing modules, which are essentially error correction modules, on the same train set can barely optimize the ASR loss as the FT model has almost no errors on the train set already. This makes training TCPGen inefficient and causes overfitting. However, training TCPGen using our two novel losses effectively avoids the issue caused by the ASR loss and makes the training more effective.

\begin{table}[t!]
    \centering
    \caption{Effect of changing $\alpha$ on the WER of the NSC-Part-2 development set, and on the true acceptance rate (TAR) and false acceptance rate (FAR) of $P^{gen}_i$. The number of unbiased word errors caused by biased words (\#U-WE-B) and B-WER is also reported to study their relationship with FAR and TAR respectively.}
    \label{tab:ablation}
    
\begin{tabular}{lcccc}
\toprule 
Metrics & $\alpha=0.1$ & $\alpha=0.3$ & $\alpha=0.5$ & $\alpha=0.7$\\
\hline
WER     & 30.3        & 28.3        & 27.2        & \textbf{25.3} \\
FAR     & \textbf{0.08}         & 1.43         & 4.28         & 9.59 \\
\#U-WE-B     & \textbf{61}        & 62        & 74        & 73 \\
TAR     & 16.9        & 41.9        & 61.6        & \textbf{67.9} \\
B-WER     & 60.0        & 48.7        & 43.4        & \textbf{37.9} \\
\bottomrule
\end{tabular}

\end{table}

\subsection{Ablation Study}

We emphasize that with our novel $\ell_{gen}$ objective function, it is possible to analyze the FAR and TAR of $P^{gen}_i$ as discussed in Section \ref{sec:far}. Table \ref{tab:ablation} shows that as we increase $\alpha$, TAR and FAR gradually increase as a larger $\alpha$ encourages $P^{gen}_i$ to classify the $i$-th decoded token as a token that needs biasing. The results also show that a higher TAR causes a lower B-WER as more words that need biasing can be correctly biased, and a higher FAR causes more unbiased word error as more of the unbiased words are wrongly biased.

\section{Conclusion}

This paper proposes two novel training objectives for TCPGen to replace the traditional ASR loss. These objectives work together to reduce overfitting and enhance performance, achieving up to a 16.6\% relative WER reduction for Whisper and Qwen Audio.

\section{Acknowledgements}
This research is supported by the National Research Foundation, Singapore, under the AI Singapore Programme (AISG Award No.: AISG2-100E-2022-102). Any opinions, findings and conclusions or recommendations expressed in this material are those of the author(s) and do not reflect the views of National Research Foundation, Singapore. The computational work for this article was partially performed on resources of the National Supercomputing Centre, Singapore (https://www.nscc.sg).

\bibliographystyle{IEEEtran}
\bibliography{mybib}

\end{document}